\documentclass{article}

\usepackage{times}
\usepackage{graphicx} % more modern
\usepackage{subfigure}

\usepackage{natbib}

\usepackage{algorithm}
\usepackage{algorithmic}

\usepackage{hyperref}

\usepackage[accepted]{icml2016}

\usepackage{amsmath}
\usepackage{enumitem}
\newcommand{\be}{\begin{equation}}
\newcommand{\ee}{\end{equation}}
\newcommand{\bea}{\begin{eqnarray}}
\newcommand{\eea}{\end{eqnarray}}

\newcommand{\E}{\mathop{\sf E}}

\newcommand{\st}{\mathop{\rm s.t.}}

\newcommand{\BEAS}{\begin{eqnarray*}}
\newcommand{\EEAS}{\end{eqnarray*}}
\newcommand{\BEA}{\begin{eqnarray}}
\newcommand{\EEA}{\end{eqnarray}}
\newcommand{\BEQ}{\begin{equation}}
\newcommand{\EEQ}{\end{equation}}
\newcommand{\BEQS}{\begin{equation*}}
\newcommand{\EEQS}{\end{equation*}}
\newcommand{\BIT}{\begin{itemize}[noitemsep, topsep=0pt]}
\newcommand{\EIT}{\end{itemize}}
\newcommand{\BNUM}{\begin{enumerate}}
\newcommand{\ENUM}{\end{enumerate}}
\newcommand{\BP}{\begin{proof}}
\newcommand{\EP}{\end{proof}}

\newcommand{\BA}{\begin{array}}
\newcommand{\EA}{\end{array}}
\newcommand{\BT}{\begin{tabular}}
\newcommand{\ET}{\end{tabular}}

\icmltitlerunning{Fixed Point Quantization of Deep Convolutional Networks}

\begin{document}

\twocolumn[
\icmltitle{Fixed Point Quantization of Deep Convolutional Networks}

\icmlauthor{Darryl D. Lin}{darryl.dlin@gmail.com}
\icmladdress{Qualcomm Research, San Diego, CA 92121, USA}
\icmlauthor{Sachin S. Talathi}{talathi@gmail.com}
\icmladdress{Qualcomm Research, San Diego, CA 92121, USA}
\icmlauthor{V. Sreekanth Annapureddy}{sreekanthav@gmail.com}
\icmladdress{NetraDyne Inc., San Diego, CA 92121, USA}

\icmlkeywords{deep learning, quantization, fixed point}

\vskip 0.3in
]

\begin{abstract}
In recent years increasingly complex architectures for deep convolution networks (DCNs) have been proposed to boost the performance on image recognition tasks. However, the gains in performance have come at a cost of substantial increase in computation and model storage resources. Fixed point implementation of DCNs has the potential to alleviate some of these complexities and facilitate potential deployment on embedded hardware. In this paper, we propose a quantizer design for fixed point implementation of DCNs. We formulate and solve an optimization problem to identify optimal fixed point bit-width allocation across DCN layers. Our experiments show that in comparison to equal bit-width settings, the fixed point DCNs with optimized bit width allocation offer $>20$\% reduction in the model size without any loss in accuracy on CIFAR-10 benchmark. We also demonstrate that fine-tuning can further enhance the accuracy of fixed point DCNs beyond that of the original floating point model. In doing so, we report a new state-of-the-art fixed point performance of 6.78\% error-rate on CIFAR-10 benchmark.
\end{abstract}

\section{Introduction}

Recent advances in the development of deep convolution networks (DCNs) have led to significant progress in solving non-trivial machine learning problems involving image recognition \citep{Krizhevsky_2012} and speech recognition \citep{Deng_2013}. Over the last two years several advances in the design of DCNs \citep{Zeiler_2014, Simoyan_2014,Szegedy_2014,Chatfield_2014,He_2014,Ioffe_2015} have not only led to a further boost in achieved accuracy on image recognition tasks but also have played a crucial role as a feature generator for other machine learning tasks such as object detection \citep{Krizhevsky_2012} and localization \citep{Sermanet_2013}, semantic segmentation \citep{Girshick_2014} and image retrieval \citep{Krizhevsky_2012,Razavian_2014}. These advances have come with an added cost of computational complexity, resulting from DCN designs involving any combinations of: increasing the number of layers in the DCN \citep{Szegedy_2014,Simoyan_2014,Chatfield_2014}, increasing the number of filters per convolution layer \citep{Zeiler_2014}, decreasing stride per convolution layer \citep{Sermanet_2013, Simoyan_2014} and hybrid architectures that combine various DCN layers \citep{Szegedy_2014, He_2014, Ioffe_2015}.

While increasing computational complexity has afforded improvements in the state-of-the-art performance, the added burden of training and testing makes these networks impractical for real world applications that involve real time processing and for deployment on mobile devices or embedded hardware with limited power budget. One approach to alleviate this burden is to increase the computational power of the hardware used to deploy these networks. An alternative approach that may be cost efficient for large scale deployment is to implement DCNs in fixed point, which may offer advantages in reducing memory bandwidth, lowering power consumption and computation time as well as the storage requirements for the DCNs.

In general, there are two approaches to designing a fixed point DCN: (1) convert a pre-trained floating point DCN model into a fixed point model without training, and (2) train a DCN model with fixed point constraint. While the second approach may produce networks with superior accuracy numbers \citep{Rastegari_2016, Lin_2016a}, it requires tight integration between the network design, training and implementation, which is not always feasible. In this paper, we will mainly focus on the former approach. In many real-world applications a pre-trained DCN is used as a feature extractor, followed by a domain specific classifier or a regressor. In these applications, the user does not have access to the original training data and the training framework. For these types of use cases, our proposed algorithm will offer an optimized method to convert any off-the-shelf pre-trained DCN model for efficient run time in fixed point.

The paper is organized as follows:  In Section \ref{sec:related_work}, we present a literature survey of the related works. In Section \ref{sec:fixed_point_conversion}, we develop quantizer design for fixed point DCNs.  In Section \ref{sec:bitwidth_optimization} we formulate an optimization problem to identify optimal fixed point bit-width allocation per layer of DCNs to maximize the achieved reduction in complexity relative to the  loss in the classification accuracy of the DCN model. Results from our experiments are reported in Section \ref{sec:experiments} followed by conclusions in the last section.

\section{Related work}
\label{sec:related_work}

Fixed point implementation of DCNs has been explored in earlier works \citep{courbariaux2014low,gupta2015deep}. These works primarily focused on training DCNs using low precision fixed-point arithmetic. More recently, \citet{Zhouhanlin_2014} showed that deep neural networks can be effectively trained using only binary weights, which in some cases can even improve classification accuracy relative to the floating point baseline. 

The works above all focused on the approach of designing the fixed point network during training. The works of \citet{Kuyeon_2014,Sajid_2014} more closely resemble our work. In \citet{Kuyeon_2014}, the authors propose a floating point to fixed point conversion algorithm for fully-connected networks. The authors used an exhaustive search strategy to identify optimal fixed point bit-width for the entire network. In a follow-up paper \citep{Sajid_2014}, the authors applied their proposed algorithm to DCN models where they analyzed the quantization sensitivity of the network for each layer and then manually decide the quantization bit-widths. Other works that are somewhat closely related are \citet{vanhoucke2011improving, gong2014compressing}. \citet{vanhoucke2011improving} quantized the weights and activations of pre-trained deep networks using 8-bit fixed-point representation to improve inference speed. \citet{gong2014compressing} on the other hand applied codebook based on scalar and vector quantization methods in order to reduce the model size.

In the spirit of \citet{Sajid_2014}, we also focus on optimizing DCN models that are pre-trained with floating point precision. However, as opposed to exhaustive search method adopted by \citet{Sajid_2014}, our objective is to convert the pre-trained DCN model into a fixed-point model using an optimization strategy based on signal-to-quantization-noise-ratio (SQNR). In doing so, we aim to improve upon the inference speed of the network and reduce storage requirements. The benefit of our approach as opposed to the brute force method is that it is grounded in a theoretical framework and offers an analytical solution for bit-width choice per layer to optimize the SQNR for the network. This offers an easier path to generalize to networks with significantly large number of layers such as the one recently proposed by \citet{He_2015}.

Other approaches to handle complexity of deep networks include: (a) leveraging high complexity networks to boost performance of low complexity networks, as proposed in \citet{Hinton_2014}, (b) compressing neural networks using hashing \citep{Chen_2015}, and (c) combining pruning and quantization during training to reduce the model size without affecting the accuracy \citep{han2015deep}. These methods are complementary to our proposed approach and the resulting networks with reduced complexity can be easily converted to fixed point using our proposed method. In fact, the DCN model that we performed experiments with and report results on, (see Section \ref{sec:alexnet}), was trained under the dark knowledge framework by using the inception network \citep{Ioffe_2015} trained on ImageNet as the master network.

\section{Floating point to fixed point conversion}
\label{sec:fixed_point_conversion}

In this section, we will propose an algorithm to convert a floating point DCN to fixed point. For a given layer of DCN the goal of conversion is to represent the input activations, the output activations, and the parameters of that layer in fixed point. This can be seen as a process of quantization.

\subsection{Optimal uniform quantizer}
\label{sec:quantizer}

There are three inter-dependent parameters to determine for the fixed point representation of a floating point DCN: bit-width, step-size (resolution), and dynamic range. These are related as:
\BEQ\label{eq:stepsize_bitwidth}
    {\rm Range} \approx {\rm Stepsize} \cdot 2^{\rm Bitwidth}
\EEQ

Given a fixed bit-width, the trade-off is between having large enough range to reduce the chance of overflow and small enough resolution to reduce the quantization error. The problem of striking the best trade-off between overflow error and quantization error has been extensively studied in the literature. Table 1 below shows the step sizes of the optimal symmetric uniform quantizer for uniform, Gaussian, Laplacian and Gamma distributions. The quantizers are optimal in the sense of minimizing the SQNR.
\begin{table}[htb]
\caption{Step-sizes of optimal symmetric uniform quantizer for various input distributions \citep{Shi08}}
\small
\label{tab:step_size}
\begin{center}
\begin{tabular}{|c|cccc|}
\hline
{Bit-width $\beta$}  &{Uniform}  &{Gaussian}  &{Laplacian}  &{Gamma}\\
\hline
\hline
1   &1.0 	&1.596  &1.414  &1.154\\
2   &0.5 	&0.996  &1.087  &1.060\\
3   &0.25 	&0.586  &0.731  &0.796\\
4   &0.125 &0.335  &0.456  &0.540\\
\hline
\end{tabular}
\end{center}
\end{table}

For example, suppose the input is Gaussian distributed with zero mean and unit variance. If we need a uniform quantizer with bit-width of 1 (i.e. 2 levels), the best approach is to place the quantized values at -0.798 and 0.798. In other words, the step size is 1.596. If we need a quantizer with bit-width of 2 (i.e. 4 levels), the best approach is to place the quantized values at -1.494, -0.498, 0.498, and 1.494. In other words, the step size is 0.996.

In practice, however, even though a symmetric quantizer is optimal for a symmetric input distribution, sometimes it is desirable to have 0 as one of the quantized values because of the potential savings in model storage and computational complexity. This means that for a quantizer with 4 levels, the quantized values could be -0.996, 0.0, 0.996, and 1.992.

\begin{figure}[hbt]
\begin{center}
%\hspace{-0.1in}%
   \includegraphics[width=0.9\linewidth]{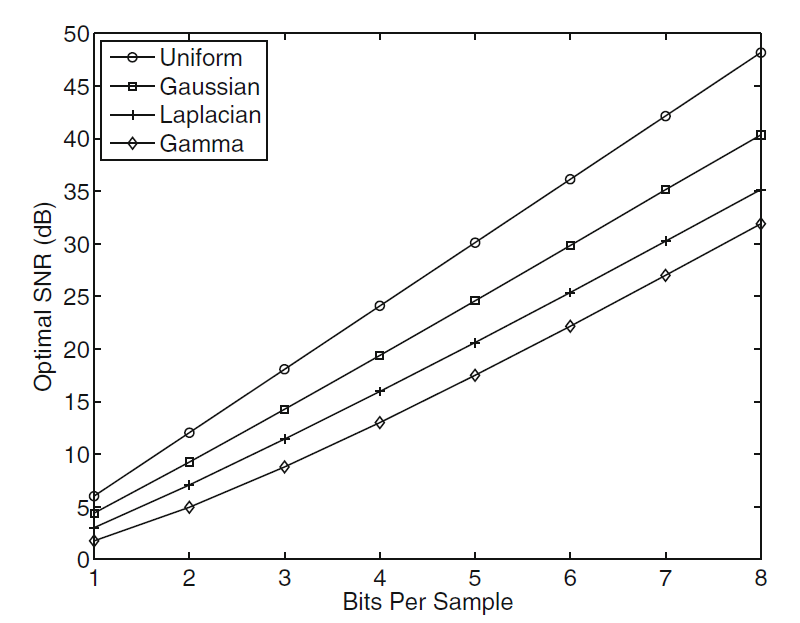}
\caption{Optimal SQNR achieved by uniform quantizer for uniform, Gaussian, Laplacian and Gamma distributions \citep{You10}} \label{fig:quant_eff}
\end{center}
\end{figure}

Assuming an optimal uniform quantizer with ideal input, the resulting SQNR as a function of the bit-width is shown in Figure \ref{fig:quant_eff}. It can be observed that the quantization efficiency decreases as the Kurtosis of the input distribution increases.

Another take-away from this figure is that there is an approximately linear relationship between the bit-width and resulting SQNR:
\BEQ\label{eq:kappa_beta}
    \gamma_{\rm dB} \approx \kappa \cdot \beta
\EEQ
where $\gamma_{\rm dB}=10\log_{10}(\gamma)$ is the SQNR in dB, $\kappa$ is the quantization efficiency, and $\beta$ is the bit-width. Note that the slopes of the lines in Figure \ref{fig:quant_eff} depict the optimal quantization efficiency for ideal distributions. The quantization efficiency for uniform distribution is the well-known value of 6dB/bit \citep{Shi08}, while the quantization efficiency for Gaussian distribution is about 5dB/bit \citep{You10} . Actual quantization efficiency for non-ideal inputs can be significantly lower. Our experiments show that the SQNR resulting from uniform quantization of the actual weights and activations in the DCN is between 2 to 4dB/bit.

\subsection{Empirical distributions in a pre-trained DCN}

\begin{figure}[htbp]
  \centering
  \subfigure[Histogram of weights]{\includegraphics[width=0.45\linewidth]{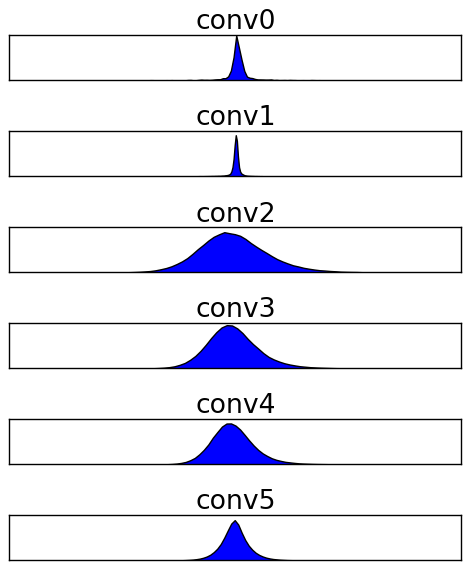}\label{fig:weights_pdf}}
  \qquad
  \subfigure[Histogram of activations]{\includegraphics[width=0.45\linewidth]{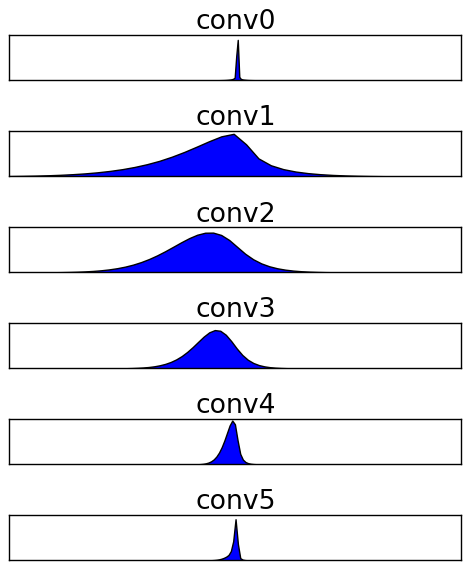}\label{fig:acts_pdf}}
  \caption{Distribution of weights \& activations in a DCN design for CIFAR-10 benchmark.}
\end{figure}

Figure \ref{fig:weights_pdf} and Figure \ref{fig:acts_pdf} depict the empirical distributions of weights and activations, respectively for the convolutional layers of the DCN we designed for CIFAR-10 benchmark (see Section \ref{sec:experiments}). Note that the activations plotted here are before applying the activation functions.

Given the similarity of these distributions to the Gaussian distribution, in all our analysis we have assumed Gaussian distribution for both weights and activations. However, we also note that the distribution of weights and activations for some layers are less Gaussian-like. It will therefore be of interest to experiment with step-sizes for other distributions (see Table \ref{tab:step_size}), which is beyond the scope of present work.

\subsection{Model conversion}

Any floating point DCN model can be converted to fixed point by following these steps:
\BIT
	\item Run a forward pass in floating point using a large set of typical inputs and record the activations.
	\item Collect the statistics of weights, biases and activations for each layer.
	\item Determine the fixed point formats of the weights, biases and activations for each layer.
\EIT

Note that determining the fixed point format is equivalent to determining the resolution, which in turn means identifying the number of fractional bits it requires to represent the number. The following equations can be used to compute the number of fractional bits:
\BIT
    \item Determine the effective standard deviation of the quantity being quantized: $\xi$.
    \item Calculate step size via Table \ref{tab:step_size}: $\displaystyle s = \xi \cdot {\rm Stepsize}(\beta)$.
    \item Compute number of fractional bits: $\displaystyle n = -\lceil \log_2 s \rceil$.
\EIT
In these equations,
\BIT
    \item $\xi$ is the effective standard deviation of the quantity being quantized, an indication of the width of the distribution we want to quantize. For example, if the quantized quantities follow an ideal zero mean Gaussian distribution, then $\xi=\sigma$, where $\sigma$ is the true standard deviation of quantized values. If the actual distribution has longer tails than Gaussian, which is often the case as shown in Figure \ref{fig:weights_pdf} and \ref{fig:acts_pdf}, then $\xi>\sigma$. In our experiments in Section \ref{sec:experiments}, we set $\xi=3\sigma$.
    \item ${\rm Stepsize}(\beta)$ is the optimal step size corresponding to quantization bit-width of $\beta$, as listed in Table \ref{tab:step_size}.
%    \item $s$ is the computed step size for the quantized distribution.
    \item $n$ is the number of fractional bits in the fixed point representation. Equivalently, $2^{-n}$ is the resolution of the fixed point representation and a quantized version of $s$. Note that $\lceil \cdot \rceil$ is one choice of a rounding function and is not unique.
\EIT

\section{Bit-width optimization across a deep network}
\label{sec:bitwidth_optimization}

In the absence of model fine-tuning, converting a floating point deep network into a fixed point deep network is essentially a process of introducing quantization noise into the neural network. It is well understood in fields like audio processing or digital communications that as the quantization noise increases, the system performance degrades. The effect of quantization can be accurately captured in a single quantity, the SQNR.

In deep learning, there is not a well-formulated relationship between SQNR and classification accuracy. However, it is reasonable to assume that in general higher quantization noise level leads to worse classification performance. Given that SQNR can be approximated theoretically and analyzed layer by layer, we focus on developing a theoretical framework to optimize for the SQNR. We then conduct empirical investigations into how the proposed optimization for SQNR affect classification accuracy of the DCN.

%While the relationship between SQNR and classification accuracy is not exactly monotonic, as our experiments in Section \ref{sec:sqnr_validation} shows, it is a useful approximation. This approximation is also important because it is usually difficult to directly predict the classification accuracy. SQNR, on the other hand, can be approximated theoretically and analyzed layer by layer, as can be seen in the next section.

\subsection{Impact of quantization on SQNR}
\label{sec:quantization_sqnr}

In this section, we will derive the relationship between the quantization of the weight, bias and activation values respectively, and the resulting output SQNR.
\subsubsection{Quantization of individual values}

Quantization of individual values in a DCN, whether it is an activation or weight value, readily follows the quantizer discussion in Section \ref{sec:quantizer}. For instance, for weight value $w$, the quantized version $\tilde{w}$ can be written as:
\BEQ
    \tilde{w} = w + n_w,
\EEQ
where $n_w$ is the quantization noise. As illustrated in Figure \ref{fig:quant_eff}, if $w$ approximately follows a uniform, Gaussian, Laplacian or Gamma distribution, the SQNR, $\gamma_w$, as a result of the quantization process can be written as:
\BEQ \label{eq:gamma_w}
    10\log (\gamma_w) = 10\log \frac{\E[w^2]}{\E[n_w^2]} \approx \kappa \cdot \beta,
\EEQ
where $\kappa$ is the quantization efficiency and $\beta$ is the quantizer bit-width.

\subsubsection{Quantization of both activations and weights}

Consider the case where weight $w$ is multiplied by activation $a$, where both $w$ and $a$ are quantized with quantization noise $n_w$ and $n_a$, respectively. The product can be approximated, for small $n_w$ and $n_a$, as follows:
\BEQ\BA{rcl}
    \tilde{w}\cdot \tilde{a} &=& (w + n_w) \cdot (a + n_a)\\
    &=& w\cdot a + w\cdot n_a + n_w\cdot a + n_w\cdot n_a \\
    &\cong& w\cdot a + w\cdot n_a + n_w\cdot a.
\EA\EEQ
The last equality holds if $|n_a|<<|a|$ and $|n_w|<<|w|$. A very important observation is that the SQNR of the product, $w\cdot a$, as a result of quantization, satisfies
\BEQ\label{eq:gamma_w_a}
    \frac{1}{\gamma_{w\cdot a}} = \frac{1}{\gamma_{w}} + \frac{1}{\gamma_{a}}.
\EEQ
This is characteristic of a linear system. The defining benefit of this realization is that introducing quantization noise to weights and activations independently is equivalent to adding the total noise after the product operation in a normalized system. This property will be used in later analysis.

\subsubsection{Forward pass through one layer}

In a DCN with multiple layers, computation of the $i$th activation in layer $l+1$ of the DCN can be expressed as follows:
\BEQ\label{eq:a_i}
    a_i^{(l+1)}=\sum_{j=1}^N w_{i,j}^{(l+1)}a_j^{(l)} + b_i^{(l+1)},
\EEQ
where $(l)$ represents the $l$th layer, $N$ represents number of additions, $w_{i,j}$ represents the weight and $b_i$ represents the bias.

Ignoring the bias term for the time being, since $a_i^{(i+1)}$ is simply a sum of terms like $w_{i,j}^{(l+1)}a_j^{(l)}$, which when quantized all have the same SQNR $\gamma_{w^{(l+1)}\cdot a^{(l)}}$. Assuming the product terms $w_{i,j}^{(l+1)}a_j^{(l)}$ are independent, it follows that the value of $a_i^{(i+1)}$, before further quantization, has inverse SQNR that equals
\BEQ\label{eq:gamma_a}
    \frac{1}{\gamma_{w_{i,j}^{(l+1)} a_j^{(l)}}} = \frac{1}{\gamma_{w_{i,j}^{(l+1)}}} + \frac{1}{\gamma_{a_j^{(l)}}} = \frac{1}{\gamma_{w^{(l+1)}}} + \frac{1}{\gamma_{a^{(l)}}}
\EEQ
After $a_i^{(l+1)}$ is quantized to the assigned bit-width, the resulting inverse SQNR then becomes $\frac{1}{\gamma_{a^{(l+1)}}} + \frac{1}{\gamma_{w^{(l+1)}}} + \frac{1}{\gamma_{a^{(l)}}}$.
We are not considering the biases in this analysis because, assuming that the biases are quantized at the same bit-width as the weights, the SQNR is dominated by the product term $w_{i,j}^{(l+1)}a_j^{(l)}$. Note that Equation \ref{eq:gamma_a} matches rather well with experiments as shown in Section \ref{sec:sqnr_validation}, even though the independence assumption of $w_{i,j}^{(l+1)}a_j^{(l)}$ does not always hold.

\subsubsection{Forward pass through entire network}

Equation \ref{eq:gamma_a} can be generalized to all the layers in a DCN (although we have found empirically that the approximation applies better for convolutional layers than fully-connected layers). Consider layer $L$ in a deep network, where all the activations and weights are quantized. Extending Equation \ref{eq:gamma_a}, we obtain the SQNR ($\gamma_{\rm output}$) at the output of layer $L$ as:
\BEQ\label{eq:gamma_output}\displaystyle
    \frac{1}{\gamma_{\rm output}} = \frac{1}{\gamma_{{a}^{(0)}}} + \frac{1}{\gamma_{{w}^{(1)}}} + \frac{1}{\gamma_{{a}^{(1)}}} + \cdots + \frac{1}{\gamma_{{w}^{(L)}}} + \frac{1}{\gamma_{{a}^{(L)}}}
\EEQ

In other word, the SQNR at the output of a layer in DCN is the \emph{Harmonic Mean} of the SQNRs of all preceding quantization steps. This simple relationship reveals some very interesting insights:
\BIT
    \item All the quantization steps contribute equally to the overall SQNR of the output, regardless if it's the quantization of weights, activations, or input, and irrespective of where it happens (at the top or bottom of the network).
    \item Since the output SQNR is the harmonic mean, the network performance will be dominated by the worst quantization step. For example, if the activations of a particular layer has a much smaller bit-width than other layers, it will be the bottleneck of network performance, because based on Equation \ref{eq:gamma_output}, $\gamma_{\rm output} \leq \gamma_{a^{(l)}}$ for all $l$.
    \item Depth makes quantization more challenging, but not exceedingly so. The rest being the same, doubling the depth of a DCN will half the output SQNR (3dB loss). But this loss can be readily recovered by adding $1$ bit to the bit-width of all weights and activations, assuming the quantization efficiency is more than 3dB/bit. However, this theoretical prediction will need to be empirically verified in future works.
\EIT

\subsubsection{Effects of other network components}
\begin{itemize}
\item {\bf Batch normalization}: Batch normalization \citep{Ioffe_2015} improves the speed of training a deep network by normalizing layer inputs. After the network is trained, the batch normalization layer is a fixed linear transformation and can be absorbed into the neighboring convolutional layer or fully-connected layer. Therefore, the quantization effect due to batch normalization does not need to be explicitly modeled.

\item {\bf ReLU}: In Equation \ref{eq:a_i}, for simplicity we omitted the activation function applied to $a_j^{(l)}$. When the activation function is ReLU and the quantization noise is small, all the positive values at the input to the activation function will have the same SQNR at the output, and the negative values become zero (effectively reducing the number of additions, $N$). In other words,
\BEQ\label{eq:a_i_relu}\BA{rcl}
    a_i^{(l+1)}&=&\sum_{j=1}^N w_{i,j}^{(l+1)}g(a_j^{(l)}) + b_i^{(l+1)}\\
    &=&\sum_{j=1}^M w_{i,j}^{(l+1)}a_j^{(l)} + b_i^{(l+1)},
\EA\EEQ
where $g(\cdot)$ is the ReLU function and $M\leq N$ is the number of $a_j^{(l)}$'s that are positive.

In this case, the ReLU function has little impact on the SQNR of $a_i^{(l+1)}$. ReLU only starts to affect SQNR calculation when the perturbation caused by quantization is sufficiently large to alter the sign of $a_j^{(l)}$. Therefore, our analysis may become increasingly inaccurate as the bit-width becomes too small (quantization noise too large).

\item {\bf Non-ReLU activations}: Other nonlinear activation functions such as tanh, sigmoid, PReLU functions are much harder to model and analyze. However, in Section \ref{sec:cross_layer_opt} we will see that applying the analysis in this section to a network with PReLU activation functions still yields useful enhancements.

\end{itemize}

\subsection{Cross-layer bit-width optimization}\label{sec:opt_bitwidth_derivation}

From Equation \ref{eq:gamma_output}, it is seen that trade-offs can be made between quantizers of different layers to produce the same $\gamma_{\rm output}$. That is to say, we can choose to use smaller bit-widths for some layers by increasing bit-widths for other layers. For example, this may be desirable because of the following reasons:
\BIT
    \item Some layers may require a large number of computations (multiply-accumulate operations). Reducing the bit-widths for these layers would reduce the overall network computation load.
    \item Some layers may contain a large number of network parameters (weights). Reducing the weight bit-widths for these layers would reduce the overall model size.
\EIT

Interestingly, such objectives can be formulated as an optimization problem and solved exactly. Suppose our goal is to reduce model size while maintaining a minimum SQNR at the DCN output. We can use $\rho_i$ as the scaling factor at quantization step $i$, which in this case represents the number of parameters being quantized in the quantization step. The problem can be written as:
\BEQ\BA{rl}
    \min_{\gamma_i}  \, \sum_i \rho_i \left(\dfrac{10\log \gamma_i}{\kappa}\right), &\st \, \sum_i \dfrac{1}{\gamma_i} \leq \dfrac{1}{\gamma_{\rm min}}
\EA\EEQ

where $10\log \gamma_i$ is the SQNR in dB domain, and $(10\log \gamma_i)/\kappa$ is the bit-width in the $i$th quantization step according to Equation \ref{eq:kappa_beta}.  $\gamma_{\rm min}$ is the minimum output SQNR required to achieve a certain level of accuracy. The optimization constraint follows from Equation \ref{eq:gamma_output} that the output SQNR is the harmonic mean of the SQNR of intermediate quantization steps.

Substituting by $\lambda_i=\dfrac{1}{\gamma_i}$ and removing the constant scalars from the objective function, the problem can be reformulated as:
%\BEQ\BA{rl}
%    \min & -\dfrac{\rho_i}{\kappa} \sum_i 10\log \lambda_i\\
%    \st & \sum_i \lambda_i \geq {\rm \gamma_{min}}
%\EA\EEQ
\BEA\BA{rl}
    \min_{\lambda_i} \, -\sum_i \rho_i \log \lambda_i, &\st \, \sum_i \lambda_i \leq C
\EA\EEA
where the constant $C= \dfrac{1}{\rm \gamma_{min}}$. This is a classic convex optimization problem with the \emph{water-filling} solution \citep{Boyd_2004}: $\dfrac{\rho_i}{\lambda_i}={\rm constant}$.

Or equivalently,
\BEQ
    \rho_i \gamma_i = {\rm constant}
\EEQ
Recognizing that $10\log \gamma_i = \kappa \beta_i$ based on Equation \ref{eq:kappa_beta}, the solution can be rewritten as:
\BEQ
    \frac{10\log \rho_i}{\kappa} + \beta_i = {\rm constant}
\EEQ
In other words, the difference between the optimal bit-widths of two quantization steps is inversely proportional to the difference of $\rho$'s in dB, scaled by the quantization efficiency.
\BEQ
    \beta_i-\beta_j = \dfrac{10\log (\rho_j/\rho_i)}{\kappa}
\EEQ
This is a surprisingly simple and insightful relationship. For example, assuming $\kappa=3$dB/bit, the bit-widths $\beta_i$ and $\beta_j$ would differ by 1 bit if $\rho_j$ is 3dB (or 2x) larger than $\rho_i$. More specifically, for model size reduction, layers with more parameters should use relatively lower bit-width, as it leads to better model compression under the overall SQNR constraint.

\section{Experiments}
\label{sec:experiments}

In this section we study the effect of reduced bit-width for both weights and activations versus traditional 32-bit single-precision floating point approach (16-bit half-precision floating point is expected to produce comparable results as single-precision in most cases). In particular, we will implement the fixed point quantization algorithm described in Section \ref{sec:fixed_point_conversion} and investigate the effectiveness of the bit-width optimization algorithm in Section \ref{sec:bitwidth_optimization}. In addition, using the quantized fixed point network as the starting point, we will further fine-tune the fixed point network within the restricted alphabets of weights and activations.

\subsection{Bit-width optimization for CIFAR-10 classification}
\label{sec:cross_layer_opt}

We evaluate our proposed cross-layer bit-width optimization algorithm on the CIFAR-10 benchmark using the algorithm prescribed in Section \ref{sec:opt_bitwidth_derivation}.

\begin{table}[htb]
\caption{Parameters per layer in our CIFAR-10 network}
\small
\label{tab:model_details}
\begin{center}
\begin{tabular}{|c|c|c|c|c|}
%\multicolumn{1}{c|}{Layer}  &\multicolumn{1}{c}{Input channels}  &\multicolumn{1}{c}{Output channels}  &\multicolumn{1}{c}{Output map dimension}  &\multicolumn{1}{c}{Filter dimension}  &\multicolumn{1}{c}{No. params (Mill)}\\
\hline
Layer   &\parbox[t]{1.3cm}{\centering Input\\channels}  &\parbox[t]{1.3cm}{\centering Output\\img size}  &\parbox[t]{1.3cm}{\centering Filter\\dim}   &\parbox[t]{1.3cm}{\centering Params\\($\times 10^6$)}\\
\hline
\hline
conv0   &3   	&22$\times$22 	&3$\times$3 	&0.007\\
conv1   &256 	&12$\times$12 	&3$\times$3 	&0.295\\
conv2   &128 	&10$\times$10 	&3$\times$3 	&0.295\\
conv3   &256 	&8$\times$8  	&3$\times$3 	&0.590\\
conv4   &256 	&5$\times$5  	&3$\times$3 	&0.590\\
conv5   &256 	&2$\times$2  	&7$\times$7 	&1.606\\
fc0        &128 	&-  			&- 			&0.005\\
\hline
\end{tabular}
\end{center}
\end{table}

In Table \ref{tab:model_details}, we compute the number of parameters in each layer of our CIFAR-10 network. Consider the objective of minimizing the overall model size. Provided that the quantization efficiency $\kappa=3$dB/bit, our derivation in Section \ref{sec:opt_bitwidth_derivation} shows that the optimal bit-width of layer conv0 and conv1 would differ by $10\log (0.295/0.007)/\kappa = 5$bits. Similarly, assuming the bit-width for layer conv0 is $\beta_0$, the subsequent convolutional layers should have bit-width values as indicated in Table \ref{tab:opt_bitwidth}.

\begin{table}[htb]
\caption{Optimal bit-width allocation in our CIFAR-10 network, assuming the bit-width of layer conv0 is $\beta_0$}
\small
\label{tab:opt_bitwidth}
\begin{center}
\begin{tabular}{c|cccccc}
\hline
Layer&\parbox[t]{0.8cm}{conv1}   &\parbox[t]{0.8cm}{conv2} &\parbox[t]{0.8cm}{conv3}   &\parbox[t]{0.8cm}{conv4} &\parbox[t]{0.8cm}{conv5}\\
\hline
Bit-width &$\beta_0-5$   &$\beta_0-5$   &$\beta_0-6$  &$\beta_0-6$   &$\beta_0-8$\\
\hline
\end{tabular}
\end{center}
\end{table}

In our experiment in this section, we will ignore the fully-connected layer and assume a fixed bit-width of 16 for both weights and activations. This is because fully-connected layers have different SQNR characteristics and need to be optimized separately. Although the fully-connected layers can often be quantized more aggressively than convolutional layers, since the number of parameters of fc0 is very small in this experiment, we will set the bit-width to a large value to eliminate the impact of quantizing fc0 from the analysis, knowing that the large bit-width of fc0 has very little impact on the overall model size. We will also set the activation bit-widths of all the layers to a large value of 16 because they do not affect the model size.

\begin{figure}[htb]\small
\begin{center}
    \includegraphics[width=0.8\linewidth, bb=110 230 480 535]{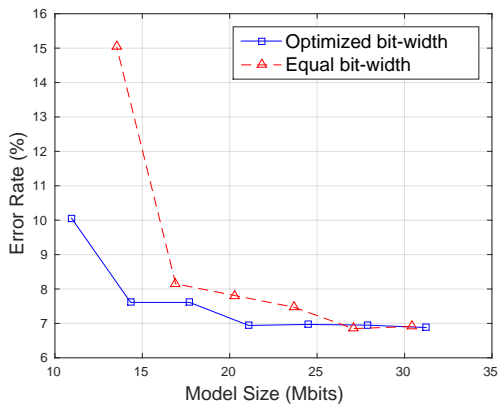}
 %  \vspace{1.9in}
\caption{Model size vs. error rate with and without cross-layer bit-width optimization (CIFAR-10)} \label{fig:equal_vs_opt}
\end{center}
\end{figure}

Figure \ref{fig:equal_vs_opt} displays the model size vs. error rate in a scatter plot, we can clearly see the advantage of cross-layer bit-width optimization. When the model size is large (bit-width is high), the error rate saturates at around 6.9\%. When the model size reduces below approximately $25$Mbits, the error rate starts to increase quickly as the model size decreases. In this region, cross-layer bit-width optimization offers $>20\%$ reduction in model size for the same performance.

\subsection{Bit-width optimization for ImageNet classification}
\label{sec:alexnet}

In Section \ref{sec:cross_layer_opt}, we performed cross-layer bit-width optimization for our CIFAR-10 network with the objective of minimizing the overall model size while maintaining accuracy. Here we carry out a similar exercise for an AlexNet-like DCN that is trained on ImageNet-1000. The DCN architecture is described in Table \ref{tab:model_details_alexnet}.

\begin{table}[htb]
\caption{Parameters per layer in our AlexNet implementation}
\small
\label{tab:model_details_alexnet}
\begin{center}
\begin{tabular}{|c|c|c|c|c|c}
%\multicolumn{1}{c|}{Layer}  &\multicolumn{1}{c}{Input channels}  &\multicolumn{1}{c}{Output channels}  &\multicolumn{1}{c}{Output map dimension}  &\multicolumn{1}{c}{Filter dimension}  &\multicolumn{1}{c}{No. params (Mill)}\\
\hline
Layer   &\parbox[t]{1.3cm}{\centering Input\\channels}    &\parbox[t]{1.3cm}{\centering Output\\img size}  &\parbox[t]{1.3cm}{\centering Filter\\dim}   &\parbox[t]{1.3cm}{\centering Params\\($\times 10^6$)}\\
\hline
\hline
conv1   &3      	&112$\times$ 112  	&7$\times$7 	&0.014\\
conv2   &96     	&28$\times$ 28    	&5$\times$5 	&0.384\\
conv3   &160    &14$\times$ 14    	&3$\times$3 	&0.277\\
conv4   &192    &14$\times$ 14    	&3$\times$3 	&0.332\\
conv5   &192    &14$\times$ 14    	&3$\times$3 	&0.277\\
fc1        &160    &-              		&- 			&16.056\\
fc2        &2048  &-              		&- 			&4.194\\
\hline
\end{tabular}
\end{center}
\end{table}

For setting the bit-width of convolutional layers of this DCN, we follow the steps in Section \ref{sec:cross_layer_opt} with the assumption that the bit-width for layer conv1 is $\beta_1$. The resulting bit-width allocation for all convolutional layers is summarized in Table \ref{tab:opt_bitwidth_alexnet}.

\begin{table}[htb]
\caption{Optimal bit-width allocation in our AlexNet-like network, assuming the bit-width of layer conv1 is $\beta_1$}
\small
\label{tab:opt_bitwidth_alexnet}
\begin{center}
\begin{tabular}{c|ccccc}
\hline
Layer &conv2 &conv3   &conv4 &conv5\\
\hline
Bit-width &$\beta_1-5$   &$\beta_1-4$   &$\beta_1-5$  &$\beta_1-4$ \\
\hline
\end{tabular}
\end{center}
\end{table}

For fully-connected layers we first keep the network as floating point and quantize the weights of fully-connected layers only. We then reduce bit-width of fully-connected layers until the classification accuracy starts to degrade. We found that the minimum bit-width for the fully-connected layers before performance degradation occurs is 6.

\begin{figure}[htb]\small
\begin{center}
   \includegraphics[width=0.8\linewidth, bb=110 230 480 535]{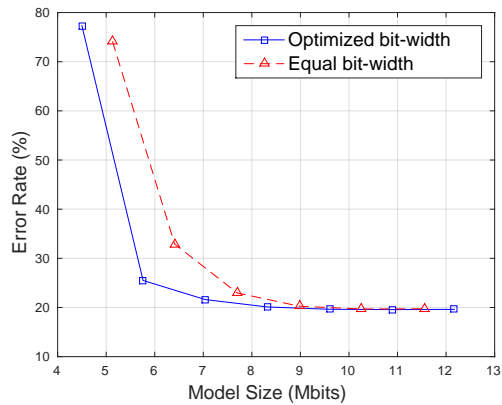}
 %  \vspace{1.9in}
\caption{Model size vs. top-5 error rate with and without cross-layer bit-width optimization (ImageNet)} \label{fig:equal_vs_opt_alexnet}
\end{center}
\end{figure}

Figure \ref{fig:equal_vs_opt_alexnet} depicts the convolutional layer model size vs. top-5 error rate tradeoff for both optimized bit-width and equal bit-width scenarios. Similar to our CIFAR-10 network, there is a clear benefit of cross-layer bit-width optimization in terms of model size reduction. In some regions the saving can be upto 1Mbits.

However, unlike our CIFAR-10 network where the convolutional layers make up most of the model size, in AlexNet-like DCN the fully-connected layers dominate in terms of number of parameters. With bit-width of 6, the overall size of fully-connected layers is more than 100Mbits. This means that the saving of 1Mbits brings in less than 1\% of overall model size reduction! This is in clear contrast to the 20\% model size reduction we reported for our CIFAR-10 network. Therefore, it is worth nothing that the proposed cross-layer layer bit-width optimization algorithm is most effective when the network size is dominated by convolutional layers, and is less effective otherwise.

The experiments on our CIFAR-10 network and AlexNet-like network demonstrate that the proposed cross-layer bit-width optimization offers clear advantage over the equal bit-width scheme. As summarized in Table \ref{tab:opt_bitwidth} and Table \ref{tab:opt_bitwidth_alexnet}, a simple offline computation of the inter-layer bit-width relationship is all that is required to perform the optimization.

However, in the absence of customized design, the implementation of the optimized bit-widths can be limited by the software or hardware platform on which the DCN operates. Often, the optimized bit-width needs to be rounded up to the next supported bit-width, which may in turn impact the network classification accuracy and model size.

\subsection{Validation for SQNR prediction}
\label{sec:sqnr_validation}
To verify that our SQNR calculation presented in Section \ref{sec:quantization_sqnr} is valid, we will conduct a small experiment. More specifically, we will focus on the optimized networks in Figure \ref{fig:equal_vs_opt} and compare the measured SQNR per layer to the SQNR predictions according to Equation \ref{eq:gamma_w} and \ref{eq:gamma_a}.
\begin{table}[h]
\caption{Predicated SQNR vs. measured SQNR (in dB) in our CIFAR-10 network}
\small
\label{tab:sqnr_compare}
\begin{center}
\begin{tabular}{|c|cc|cc|cc|cc}
\hline
\multicolumn{1}{|c|}{}  &\multicolumn{2}{c|}{Example 1} &\multicolumn{2}{c|}{Example 2} \\
\hline
&\parbox[t]{1.2cm}{\centering Predicted} &\parbox[t]{1.2cm}{\centering Measured} &\parbox[t]{1.2cm}{\centering Predicted} &\parbox[t]{1.2cm}{\centering Measured} \\
\hline
\hline
conv1       &23.83  &24.93	&20.85	&20.12\\
conv2       &20.9	&23.75	&17.91	&16.82\\
conv3       &17.93	&22.62	&14.94	&17.83\\
conv4       &16.16	&21.81	&13.2	&14.74\\
conv5       &12.54	&18.01	&9.55	&9.26 \\
\hline
\end{tabular}
\end{center}
\end{table}

Table \ref{tab:sqnr_compare} contains the comparison between the theoretical SQNR and the measured SQNR (in dB) for layers conv1 to conv5 for two of the optimized networks. We observe that while the two SQNR values do not match numerically, they follow similar decreasing trend as the activations propagate deeper into the network. It should be noted that our theoretical SQNR predictions are based purely on  the weight and activation bit-widths of each layer as well as the quantization efficiency $\kappa$. The theory does not rely on any information related to the network parameters or the data it is tested on.

\subsection{Model fine-tuning}\label{sec:fine_tuning_results}

Although our focus is fixed point implementation without training, our quantizer design can also be used as a starting point for further model fine-tuning when the training model and training parameters are available.

\begin{table}[htb]
\caption{CIFAR-10 classification error rate with different bit-width combinations}
\small
\label{tab:results_no_ft}
\begin{center}
\begin{tabular}{|c|cccc|}
\hline
\multicolumn{1}{|c|}{Activation}  &\multicolumn{4}{c|}{Weight Bit-width} \\
\multicolumn{1}{|c|}{Bit-width} &\multicolumn{1}{c}{4}  &\multicolumn{1}{c}{8}  &\multicolumn{1}{c}{16}  &\multicolumn{1}{c|}{Float}\\
\hline
\hline
4       &8.30  &7.50  &7.40  &7.44\\
8       &7.58  &6.95  &6.95  &6.78\\
16      &7.58  &6.82  &6.92  &6.83\\
Float   &7.62  &6.94  &6.96  &6.98\\
\hline
\end{tabular}
\end{center}
\end{table}

Table \ref{tab:results_no_ft} contains the classification error rate (in \%) for the CIFAR-10 network after fine-tuning the model for 30 epochs. We experiment with different weight and activation bit-width combinations, ranging from floating point to 4-bit, 8-bit, and 16-bit fixed point. It is shown that even the (4b, 4b) bit-width combination works well (8.30\% error rate) when the network is fine-tuned after quantization. In addition, the (float, 8b) setting generates an error rate of 6.78\%, which is the new state-of-the-art result even though the activations are only 8-bit fixed point values. This may be attributed to the regularization effect of the added quantization noise \citep{Zhouhanlin_2014, Luo_2014}.

\section{Conclusions}

Fixed point implementation of deep networks is important for real world embedded applications, which involve real time processing with limited power budget. In this paper, we develop a principled approach to converting a pre-trained floating point DCN model to its fixed point equivalent. We show that the naive method of quantizing all the layers in the DCN with uniform bit-width value results in DCN networks with subpar performance in terms of error rates relative to our proposed approach of SQNR based optimization of bit-widths. Specifically, we present results for a floating point DCN trained CIFAR-10 benchmark, which on conversion to its fixed point counter-part results in $>$20 \% reduction in model size without any loss in accuracy. We note that our proposed algorithm facilitates easy conversion of any off-the-shelf DCN model for efficient real world on-device application. Finally, through fine-tuning experiments we demonstrate that our quantizer design methodology is a useful starting point for further model fine-tuning after the floating-point-to-fixed-point conversion.

\section*{Acknowledgements}

We would like to acknowledge fruitful discussions and valuable feedback from our colleagues: David Julian, Anthony Sarah, Daniel Fontijne, Somdeb Majumdar, Aniket Vartak, Blythe Towal, and Mark Staskauskas.

\bibliography{icml2016_conference}
\bibliographystyle{icml2016}

\end{document}